\documentclass[lettersize,journal]{IEEEtran}
\usepackage{amsmath,amsfonts}
\usepackage{algorithmic}
\usepackage{algorithm}
\usepackage{array}
\usepackage[caption=false,font=normalsize,labelfont=sf,textfont=sf]{subfig}
\usepackage{textcomp}
\usepackage{stfloats}
\usepackage{url}
\usepackage{verbatim}
\usepackage{graphicx}
\usepackage{cite}

\usepackage{multirow}
\usepackage{makecell}
\usepackage{soul}
\usepackage[para,online,flushleft]{threeparttable}
\usepackage{booktabs}
\usepackage{ulem}
\usepackage{bm}
\usepackage[inkscapelatex=false]{svg}
\usepackage{nicematrix, tikz}
\usepackage{ulem}
\usepackage[framemethod=tikz]{mdframed}
\usepackage{mathtools}
\renewcommand\hl[1]{#1} 
\newcommand{\highlight}[1]{#1}
\begin{document}
	
	\title{DBA-Fusion: Tightly Integrating Deep Dense Visual Bundle Adjustment with Multiple Sensors for Large-Scale  Localization and Mapping}
	
	\author{Yuxuan Zhou, Xingxing Li*, Shengyu Li, Xuanbin Wang, Shaoquan Feng, Yuxuan Tan
		\thanks{This work was supported in part by the National Key Research and Development Program of China (2021YFB2501102).\textit{(Corresponding author: Xingxing Li.)}}
		\thanks{The authors are with School of Geodesy and Geomatics, Wuhan University, China  (e-mail: xxli@sgg.whu.edu.cn).}
	}
	
	\markboth{Journal of \LaTeX\ Class Files,~Vol.~14, No.~8, August~2021}%
	{Shell \MakeLowercase{\textit{et al.}}: A Sample Article Using IEEEtran.cls for IEEE Journals}
	
	\IEEEpubid{ }
	
	\maketitle
	
	\begin{abstract}
		Visual simultaneous localization and mapping (VSLAM) has broad applications, with state-of-the-art methods leveraging deep neural networks for better robustness and applicability.  However, there is a lack of research in fusing these learning-based methods with multi-sensor information, which could be indispensable to push related applications to large-scale and complex scenarios. In this paper, we tightly integrate the trainable deep dense bundle adjustment (DBA) with multi-sensor information through a factor graph. In the framework, recurrent optical flow and DBA are performed among sequential images. The Hessian information  derived from DBA is fed into a generic factor graph for multi-sensor fusion, which employs a sliding window and supports probabilistic marginalization. A pipeline for visual-inertial integration is firstly developed, \hl{which provides} the minimum ability of metric-scale localization and mapping. Furthermore, other sensors (e.g., global navigation satellite system) are integrated for driftless and geo-referencing functionality. Extensive tests are conducted on both public datasets and self-collected datasets. The results validate the superior localization performance of our approach, which enables real-time dense mapping in large-scale environments. \hl{The code has been made open-source} (https://github.com/GREAT-WHU/DBA-Fusion).
	\end{abstract}
	
	
	\section{Introduction}
	\IEEEPARstart{V}{isual} simultaneous localization and mapping (VSLAM) is a pivotal technology in VR/AR and robotics applications\cite{ref1}. Recently, deep learning has significantly propelled the advancement of VSLAM \hl{on better accuracy, robustness and dense spatial perception by utilizing higher-level information\mbox{\cite{ref2,ref_deepfactor}}.} Due to the data-driven nature, the generalization of end-to-end VSLAM systems on new data domains is challenging. To address this issue, DROID-SLAM\cite{ref_droid} designs a recurrent structure integrating optical flow and dense bundle adjustment (DBA), which is end-to-end trainable and demonstrates excellent zero-shot performance.
	
	It's important to note that, a visual-only system has its inherent limitations and is vulnerable  to extreme illumination issues and dynamic scenarios. To address these challenges, a promising approach is to incorporate multiple sensors, such as inertial measurement units (IMU)\cite{ref_vins}, global navigation satellite system (GNSS)\cite{ref_giv} and wheel speed sensors (WSS)\cite{ref_vow} for optimal information fusion. Although the integration of VSLAM systems with multiple sensors has been widely studied, there is limited research when it comes to learning-based VSLAM frameworks. As model-driven methods are considered effective enough for the above mentioned sensors in most cases, the key challenge lies in how to probabilistically fuse multi-sensor state estimation with deep VSLAM methods, form a generic framework, and assess  its practical significance in terms of state estimation and mapping.
	
	\begin{figure}[!t]
		\centering
		\includegraphics[width=8.4cm]{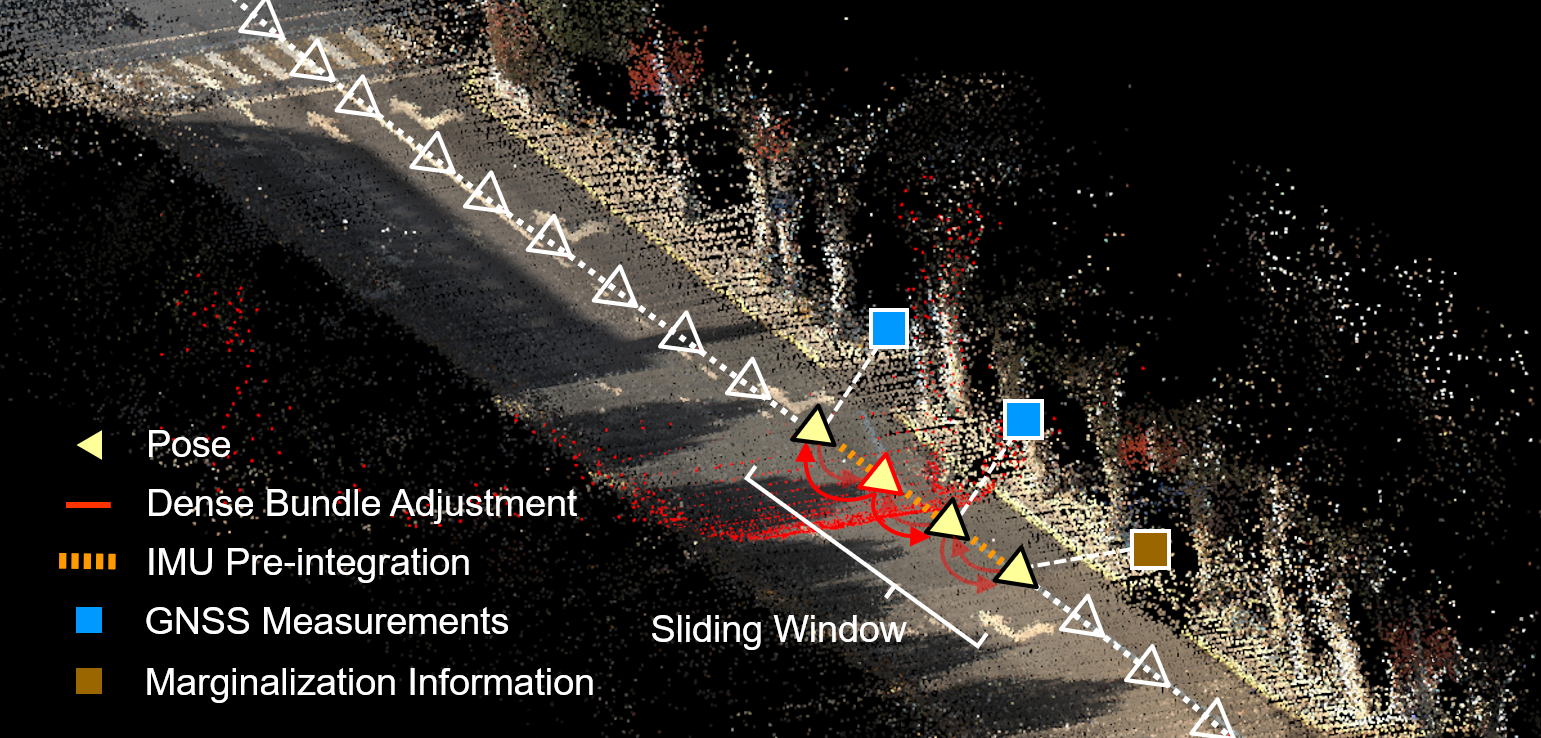}
		\caption{Illustration of the DBA-Fusion system. The pointcloud map is generated online via monucular visual-inertial-GNSS integration. For clarity, we only show 5x downsampled camera poses.}
		\label{fig_intro}
	\end{figure}

	In this work, we proposes DBA-Fusion, which provides a generic factor graph framework to tightly integrate DROID-SLAM-like deep DBA with multi-sensor information, as illustrated in Fig. \ref{fig_intro}. The system is designed to enhance the practicality of deep VSLAM through multi-sensor fusion and make it suitable for real-time large-scale  applications. The contributions of this work are summarized as follows:
	
	1) We provide a pipeline for tight visual-inertial integration which introduces the  recurrent deep DBA into a sliding-window factor graph optimization framework. 
	
	2) The proposed system supports flexible integration of multiple sensors (e.g. GNSS, WSS), thus \hl{can} be applied to large-scale, georeferenced navigation and mapping applications.
	
	3) The comprehensive performance of the  system is  evaluated on both public and self-made datasets  covering  indoor/outdoor, handheld/vehicular scenarios.
	
	\IEEEpubidadjcol

	\section{Related Work}
	\subsection{Deep  VSLAM}
	
	Deep learning has been extensively studied for integration into VSLAM in recent years with diverse motivations. Learning-based methods have been used to replace the key modules in VSLAM systems, like feature tracking\cite{ref_lg}, depth inference\cite{ref_depth} or to achieve object-level SLAM with semantic information\cite{ref_cube,ref_vdo}. There are also many works that attempt to develop end-to-end visual SLAM systems\cite{ref_e2e_vo1,ref_e2e_vo2}. Recent works have been concentrating on the neural representation of 3-D spatial map, which has shown impressive  reconstruction performance\cite{ref_coslam,ref_nicer}.
	
	Among the above methods, DROID-SLAM is a popular one which combines the advantages of traditional SLAM pipeline and end-to-end learning, and is well-known for its good generality and dense mapping capability. The core design of DROID-SLAM is a recurrent optical flow module based on gate recurrent units (GRU) and a differentiable DBA layer, which makes it decoupled with  depth inference and \hl{can} be flexibly applied to zero-shot datasets.
	
	\subsection{VSLAM with multi-sensor fusion}
	Visual-inertial integration is a commonly used scheme to overcome the limitation of visual-only SLAM, which provides scale awareness and estimation continuity with an almost minimum setup. The classic implementations could be divided into filter-based \cite{ref_msckf} and optimization-based \cite{ref_okvis}. Advanced features have been developed over the past decades, including observability constraint \cite{ref_msckf2}, direct photometric optimization\cite{ref_vidso},  map management\cite{ref_orb} and delayed marginalization\cite{ref_dmvio}. In\cite{ref_bamf}, DROID-SLAM is fused with IMU in a factor graph framework. However, marginalization and monocular setup are not considered. Recently, end-to-end visual-inertial SLAM structures have also been proposed\cite{ref_dvi}.
	
	To expand the applicability of the V-I system, many other works tend to fully utilize the information of available sensors, of which the most representative ones are wheel encoder for ground vehicles and GNSS for outdoor applications. The wheel encoder has been proved to improve the system stability and scale observability\cite{ref_vow}\cite{ref_gv}. As to GNSS integration, both loosely coupled\cite{ref_vins_fusion} and tightly coupled\cite{ref_gvins,ref_giv} methods have been proposed, which generally elevate the system to a geo-registered one and provide global driftless capability. Nevertheless, there is a noticeable absence of an implementation that incorporates advanced learning-based VSLAM into the multi-sensor fusion framework.

	\section{System Implementation}
	
	The overall design of the framework aims to \hl{flexibly fuses} a trainable VSLAM system with multi-sensor information.
	
	For the visual frontend, a recurrent optical flow module is utilized to compute  dense pixel association for image pairs in a co-visibility graph. The re-projections are used to formulate a DBA problem. The DBA information is then tightly fused with multi-sensor  through a factor graph, where a sliding window mechanism with \hl{probabilistic marginalization} is employed.
	
	As to multi-sensor fusion, IMU is firstly considered in the factor graph to enable a minimum visual-inertial odometry (VIO). 
	Besides, to improve the system applicability in large-scale scenarios, common sensors like GNSS and WSS are integrated into the factor graph when available.

	\begin{figure}[!t]
		\centering
		\includesvg[width=8.0cm]{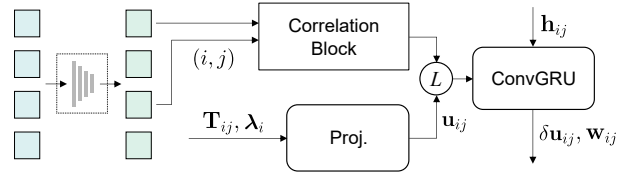}
		\caption{Illustration of the recurrent optical flow. The images are firstly encoded to feature maps. One co-visible image pair $(i,j)$ is fed into the recurrent optical flow module, as described in Sect. III-A. }
		\label{fig_of}
	\end{figure}

	\subsection{Recurrent Optical Flow}
	The recurrent optical flow applied in this work adheres to the implementation of DROID-SLAM, which is derived from the well-developed RAFT\cite{ref_raft} and tightly leverages the camera poses and depths for iterative refinement.

	\hl{The structure of the recurrent optical flow is depicted in Fig. \mbox{\ref{fig_of}}. To be specific, for a co-visible image pair (source frame $i$ projected to target frame $j$), a correlation volume is formed based on their encoded feature maps. During one update operation of the optical flow, firstly,  the  (inverse) depth map $\bm{\lambda}_{i}$ of the source frame is projected via the relative pose $\mathbf{T}_{ij}$  to estimate the initial optical flow field}
		\begin{equation}
			\highlight{		\mathbf{u}_{ij}=\Pi_c(\mathbf{T}_{ij} \circ \Pi_c^{-1} (\mathbf{u}_i,\bm{\lambda}_i)),\ \ 
				\mathbf{T}_{ij} = \mathbf{T}_j\circ \mathbf{T}_i^{-1}}
	\end{equation}
		\hl{where $\mathbf{u}_i$ is the pixel coordinates of the grid points, $\mathbf{T}_i$ and $\mathbf{T}_j$ are world-to-camera transformations, $\Pi_c$ is the camera projection model, whose inverse operation uplifts the pixel coordinates and inverse depths to 3-D points.} \hl{Notice that Eq. (1) is also used to estimate  inter-frame disparity to check the co-visibility among frames in further steps.}

	\hl{Based on the flow estimation,} a look-up operation $L$ to the correlation volume is performed. The results are then injected into the convolutional GRU with hidden state $\mathbf{h}_{ij}$ to obtain the corrected flow $\delta\mathbf{u}_{ij}$ in the residual form. Besides, the pixel-wise weight $\mathbf{w}_{ij}$ of the optical flow, which indicates geometric and appearance uncertainty, is generated to boost \hl{the downstream BA}. For real-time performance, we only consider pixels in each image with a step size of 8\hl{.} In  this case, the size $n$ of  $\bm{\lambda}_{i}$ is $ (H/8)\times(W/8)$, where $(H,W)$ is the size of the image. 
	
	The above optical flow is computed on every active image pair (termed as an edge) \hl{in the co-visibility graph}, thus to construct a DBA problem. The optimization results of DBA, i.e. the dense depths and camera poses, are recurrently fed into the optical flow module to refine everything iteratively.  Such operations would be performed continuously to adjust the optical flow until achieving a converged state, which is different from the usual forward-tracking visual frontend. The recurrent optical flow design has been verified to show good generalization on zero-shot datasets. In this work, we just use the network weights pretrained on the  synthetic TartanAir\cite{ref_tartan}  dataset for all the tests later. 

	\subsection{{Integrating DBA into a generic factor graph}}
		\hl{The basic element of DBA is one uni-directional edge in the co-visibility graph, corresponding to the re-preojection errors from one source frame to one target frame, whose linearized form is}
			\begin{equation}
				\highlight{
		\underbrace{\vphantom{\left[\setlength{\arraycolsep}{2pt}\begin{matrix}
					\bm{\xi}_i^\top & \bm{\xi}_j^\top & \left({\delta\bm{\lambda}_i}\right)_{n\times1}^\top
				\end{matrix}\right]} \left(\delta\mathbf{u}_{ij}\right)_{2n\times1} }_{\mathbf{r}}
		 =
		 	\underbrace{\vphantom{\left[\setlength{\arraycolsep}{2pt}\begin{matrix}
		 				\bm{\xi}_i^\top & \bm{\xi}_j^\top & \left({\delta\bm{\lambda}_i}\right)_{n\times1}^\top
		 			\end{matrix}\right]} \left[\setlength{\arraycolsep}{2.5pt}\begin{matrix}
			\mathbf{J}_i & \mathbf{J}_j & \mathbf{J}_{\bm{\lambda}_i}
		\end{matrix}\right]}_\mathbf{J} \ 
		{	\underbrace{\left[\setlength{\arraycolsep}{2pt}\begin{matrix}
			\bm{\xi}_i^\top & \bm{\xi}_j^\top & \left({\delta\bm{\lambda}_i}\right)_{n\times1}^\top
		\end{matrix}\right]}_\mathbf{x}}^\top}
	\end{equation}
	\hl{where $\delta\mathbf{u}_{ij}$ is the optical flow residual outputted by the optical flow module, $\bm{\xi}_i$, $\bm{\xi}_j$ are  Lie algebras of the camera poses, $\delta\bm{\lambda}_i$ is the  error state of the  source frame's inverse depth map, $\mathbf{J}_i$, $\mathbf{J}_j$, $\mathbf{J}_{\bm{\lambda}_i}$ are Jacobians derived from (1). }
	
	\hl{Taking into account the weight $\mathbf{w}_{ij}$ of the optical flow, the Hessian form of Eq. (2) is computed following }
	\begin{equation}
		\highlight{
	\mathbf{J}^\top\mathbf{W}\mathbf{r} = \mathbf{J}^\top\mathbf{W}\mathbf{J}\mathbf{x} ,\ \ \mathbf{W} = \text{diag}(\mathbf{w}_{ij})
}
	\end{equation}
	 \hl{and leads to the following equation}

	\begin{equation}
		\highlight{
		\left[\begin{matrix}
			\mathbf{v}_{ii} \\ \mathbf{v}_{ij} \\ \mathbf{z}_{ii}
		\end{matrix}\right] = \left[\begin{matrix}
			\mathbf{B}_{ii} & \mathbf{B}_{ij} & \mathbf{E}_{ii} \\
			\mathbf{B}^\top_{ij} & \mathbf{B}_{jj} & \mathbf{E}_{ij} \\
			\mathbf{E}^\top_{ii} & \mathbf{E}^\top_{ij} & \mathbf{C}_{ii}
		\end{matrix}\right]
		\left[\begin{matrix}
			\bm{\xi}_{i} \\ \bm{\xi}_{j} \\ {\delta\bm{\lambda}_i}
		\end{matrix}\right]}
	\end{equation}
	\hl{where $\mathbf{C}_{ii}$ is a $n\times n$ diagonal matrix.}
	
\begin{figure}[!t]
		\centering
		\includesvg[width=8.0cm]{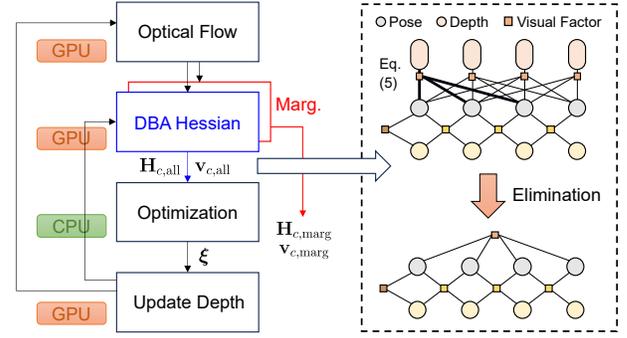}
		\caption{\hl{Workflow of DBA in the optimization framework, which employs a hierarchical iteration structure.} The right panel shows the elimination of the  depth state, \hl{which generates the Hessian $\mathbf{H}_{c,\text{all}}$,$\mathbf{v}_{c,\text{all}}$ as the visual constraint}. For marginalization, only related edges are processed to generate \hl{ $\mathbf{H}_{c,\text{marg}}$,$\mathbf{v}_{c,\text{marg}}$ }that contains the marginalized visual information.}
		\label{fig_opt}
	\end{figure}

	\hl{Considering a bundle of edges anchored on frame $i$  and projected to co-visible frames (numbered 1,\! 2, \dots,\! N as example), the  combined Hessian is constructed by positionally stacking/summing the blocks in Eq. (4) as}
	\begin{equation}
				\highlight{
		\begin{bNiceArray}[]{ccc}[margin,cell-space-limits = 1.5pt]
			\!\Sigma\mathbf{v}_{ii} \!\\
			\!	\mathbf{v}_{i1}  \!\\ 
			\!\vdots \!\\
			\!	\mathbf{v}_{iN} \Block[borders={bottom, tikz=densely dashed}]{1-1}{} \!\\
			\!	\Sigma\mathbf{z}_{ii}\!\\
		\end{bNiceArray}
		\!\!	=\!\!
		\begin{bNiceArray}[]{ccccc}[margin,cell-space-limits = 1.5pt]
			\!		\Sigma\mathbf{B}_{ii} &\! \mathbf{B}_{i1}& \!
			\cdots&\! \Block[borders={right, tikz=densely dashed}]{5-1}{}  \mathbf{B}_{iN}& \!		 \Sigma\mathbf{E}_{ii}\\
			\!		\mathbf{B}^\top_{i1} &\! \mathbf{B}_{11}&\!  &\! &\! \mathbf{E}_{i1} \\
			\!		\vdots &\! &\! \ddots &\!   &\! \vdots \\
			\Block[borders={bottom, tikz=densely dashed}]{1-5}{}
			\!				\mathbf{B}^\top_{iN} &\! &\!  &\! \mathbf{B}_{N\!N}   &\! \mathbf{E}_{iN} \\
			\!		\Sigma\mathbf{E}^\top_{ii} &\! \mathbf{E}^\top_{i1}&\! \cdots&\! \mathbf{E}^\top_{iN} &\! \Sigma\mathbf{C}_{ii} \\
		\end{bNiceArray}
		\!\!
		\begin{bNiceArray}[]{ccc}[margin,cell-space-limits = 1.5pt]
			\!	\bm{\xi}_{i}\! \\\! \bm{\xi}_{1}\! \\\! \vdots\! \\\! \bm{\xi}_{N}\!  \Block[borders={bottom, tikz=densely dashed}]{1-1}{} \\  \!{\delta\bm{\lambda}_i}\!
		\end{bNiceArray}}
	\end{equation}
	which corresponds to one visual factor depicted in Fig. \ref{fig_opt}.
	Note that no other edges in the graph are related to $\delta\bm{\lambda}_i$, thus Schur complement\cite{ref_gtsam} \hl{can} be performed on Eq. (5) to eliminate the depth state, leading to
	\begin{equation}
		\highlight{
	(\underbrace{\mathbf{B}_i - \mathbf{E}_i\mathbf{C}_i^{-1}\mathbf{E}_i^\top}_{\mathbf{H}_{c,i}}) \bm{\xi}_{i,1,\cdots,N} =  \underbrace{\mathbf{v}_i - \mathbf{E}_i \mathbf{C}_i^{-1} \mathbf{z}_i}_{\mathbf{v}_{c,i}}}
\end{equation}
	where $\mathbf{B}_i$, $\mathbf{E}_i$, $\mathbf{C}_{i}$, $\mathbf{v}_i$, $\mathbf{z}_i$ are blocks in Eq. (5) and $\mathbf{C}_{i}$ is a diagonal matrix. \hl{This actually constructs a inter-frame pose constraint (represented by $\mathbf{H}_{c,i}, \mathbf{v}_{c,i}$) which contains the linearized BA information. The calculation of Eq. (5) and Eq. (6) can be efficiently performed in parallel on GPU.
As long as the camera poses are updated, the inverse depth state  can be updated according to the bottom row of Eq. (5)}
\begin{equation}
	\delta \bm{\lambda}_i = \mathbf{C}_i^{-1} (\mathbf{z}_i-\mathbf{E}_i^\top\bm{\xi}_{i,1,\dots,N})
\end{equation}
	
	In this work, in order to tightly integrate the DBA with a generic pose-centered factor graph, \hl{all the constraints Eq. (6) derived from the co-visibility graph are stacked to form a full Hessian $\mathbf{H}_{c,\text{all}}$ and corresponding $ \mathbf{v}_{c,\text{all}}$, which is taken as a constraint among all camera poses in the factor graph. The factor graph optimization is efficiently performed on CPU,} considering that the dense depth states have been eliminated. This is actually equal to solving the full factor graph optimization in a two-step way,  similar to the implementation in\cite{ref_dmvio}.
	The pixel association (produced by the  optical flow) and the DBA formulation need to be iteratively updated for edges  in the graph, leading to the hierarchical iteration structure shown in Fig. \ref{fig_opt}.
	
	 \hl{Generally, the deep DBA provides rich geometric information (with learned uncertainty) to the factor graph, while the optimization results (poses and depths) are fed back to refine/re-weight the  optical flow recurrently.} \hl{The above pipeline also supports stereo and RGB-D setups, either by adding left-to-right edges or adding depth priors in Eq. (5).} 
	
\begin{figure}[!t]
		\centering
		\includesvg[width=8.4cm]{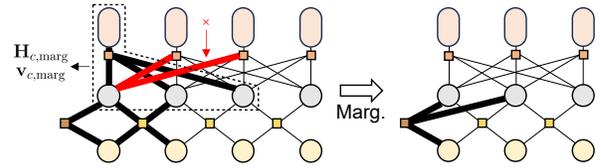}
		\caption{Marginalization of the factor graph. \hl{Visual information as well as other information related to the oldest states (black edges) is marginalized, which uses Schur complement to eliminate the oldest states and form a new constraint\mbox{\cite{ref_okvis,ref_gtsam}}.} The re-projections from newer frames to the old frame (red edges) are neglected to keep sparsity.}
		\label{fig_marg}
	\end{figure}

	Every time an old frame is slided out of the window, the camera pose and the contained depth state need to be marginalized. To keep the sparsity of the factor graph, only the re-projection residuals built upon the marginalized  depth state are considered, \hl{which forms the partial visual information $\mathbf{H}_{c,\text{marg}},\mathbf{v}_{c,\text{marg}}$ as shown in Fig. \mbox{\ref{fig_marg}}.}
	The visual information  and other information related to the oldest frame, together with the existing marginalization factor, are  used to constitute the new marginalization factor through Schur complement.

	\subsection{Multi-sensor factor graph}
	
	Applying the above design, we \hl{can} build up a generic pose-centered factor graph to solve the pose estimation, with the advantages of efficient optimization and loseless  DBA information integration. The factor graph optimization is implemented using GTSAM\cite{ref_gtsam_software}.
	
\begin{figure}[!t]
		\centering
		\includesvg[width=7.0cm]{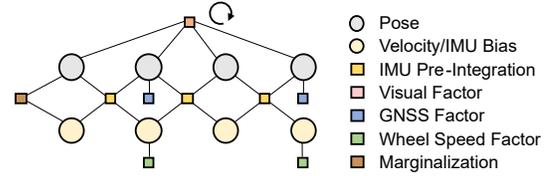}
		\caption{The generic pose-centered factor graph that is used for multi-sensor fusion. Note that the visual factor is modeled using a Hessian factor where the point depths are temporarily eliminated. The visual factor is iteratively updated with the recurrent optical flow and DBA Hessian formulation.}
		\label{fig_fg}
	\end{figure}

	Taking the commonly used sensors in vehicle application into account, the overall factor graph is depicted in Fig. \ref{fig_fg}, with the states expressed as 
	\begin{equation}
		\mathbf{x}_k = 	
		\left[\begin{matrix}
			\mathbf{T}^w_{b_k} & \mathbf{v}^w_{b_k} & \mathbf{b}_{k}
		\end{matrix}\right],  k = 1,2,3,...
	\end{equation}
	\begin{equation}
		\mathbf{T}^w_{b_k} = 	
		\left[\begin{matrix}
			\mathbf{R}^w_{b_k} & \mathbf{t}^w_{b_k} \\
			0& 1 
		\end{matrix}\right]\in SE(3),\ \ \mathbf{b}_k = \left[\begin{matrix} \mathbf{b}_{a,k} & \mathbf{b}_{g,k} \end{matrix}\right]
	\end{equation}
	where $\mathbf{T}^w_{b_k}$ is the IMU pose in the world frame, $\mathbf{v}^w_{b_k}$ is the velocity, $\mathbf{b}_{a,k}$ and $\mathbf{b}_{g,k}$ are accelerometer/gyroscope biases.

	The sensor factors are presented as follows.
	
	1) Visual constraint factor (\hl{quadratic cost function\mbox{\cite{ref_gtsam_software}}}):
	\begin{equation}
		\highlight{
		\mathbf{E}_c (\mathbf{x}_c) =\frac{1}{2} \mathbf{x}_c^\top\mathbf{H}_{c,\text{all}}\,\mathbf{x}_c - \mathbf{x}_c^\top \mathbf{v}_{c,\text{all}}}
	\end{equation}
	  where \hl{$\mathbf{H}_{c,\text{all}}$ and $\mathbf{v}_{c,\text{all}}$} are the information matrix and vector of DBA with all the depth parameters  eliminated, $ \mathbf{x}_c$ is the  Lie algebras of the camera poses $\mathbf{T}^c_w$
	\begin{equation}
		\mathbf{x}_c = 
		\left[\begin{matrix}
			{\bm{\xi}^{c_0}_w}^\top & {\bm{\xi}^{c_1}_w}^\top & \cdots & {\bm{\xi}^{c_k}_w}^\top 
		\end{matrix}\right]^\top
	\end{equation}

	In order to impose the visual constraint on the IMU-centric factor graph, the following linear transformation is employed
	\begin{equation}
		\bm{\xi}^c_w =-\text{adj}\left(\mathbf{T}^c_b\right) \bm{\xi}^w_b
	\end{equation}
	where $\bm{\xi}^c_w$ is derived from left-perturbation and $\bm{\xi}^w_b$ is derived from right-perturbation, $\mathbf{T}^c_b$ is the IMU-camera extrinsics, \hl{ $\text{adj}(\cdot)$ is the adjoint operation }.

	2) IMU preintegration factor:
	\begin{align}
		\begin{split}
			&\mathbf{r}_{b}\left( \mathbf{x}_k,\mathbf{x}_{k+1}\right)= \\
			&\left[\begin{matrix}
				{\mathbf{R}^w_{b_k}}^{\top} \left( \mathbf{p}^w_{b_{k+1}}\! -\! \mathbf{p}^w_{b_k}\!+\! \frac{1}{2}\mathbf{g}^w\Delta{t^2_k}\! -\! \mathbf{v}^w_{b_k} \Delta t_k\! -\! \Delta\tilde{\mathbf{p}}^{b_k}_{b_{k+1}} \right) \\
				{\mathbf{R}^w_{b_k}}^{\top} \left( \mathbf{v}^w_{b_{k+1}} + \mathbf{g}^w \Delta t_k - \mathbf{v}^w_{b_k}\right) - \Delta\tilde{\mathbf{v}} ^{b_k}_{b_{k+1}}  \\
				\text{Log}\left(\left({\mathbf{R}^w_{b_k}}\right)^{-1}  \mathbf{R}^w_{b_{k+1}}  {\left(\Delta\tilde{\mathbf{R}} ^{b_k}_{b_{k+1}}\right)^{-1}} \right) \\
				\mathbf{b}_{a,k+1} - \mathbf{b}_{a,k} \\
				\mathbf{b}_{g,k+1} - \mathbf{b}_{g,k}
			\end{matrix}\right]
		\end{split}
	\end{align}
	where $\Delta\tilde{\mathbf{p}}^{b_k}_{b_{k+1}}$, $\Delta\tilde{\mathbf{v}}^{b_k}_{b_{k+1}}$, $\Delta\tilde{\mathbf{R}}^{b_k}_{b_{k+1}}$ are the IMU pre-integration terms\cite{ref_imu}, $\mathbf{g}^w$ is the gravity vector, $\Delta t_k$ is the time interval.
	The integration of IMU and DBA enables the funtionality of a VIO, while certain  steps are needed to bootstrap the system.
	
	3) GNSS position factor:
	\begin{equation}
		\mathbf{r}_g (\mathbf{x}_k) =\mathbf{T}^n_w\circ \mathbf{T}^w_{b_k}\circ\mathbf{t}^b_g - \tilde{\mathbf{p}}^n_g
	\end{equation}
	where $\tilde{\mathbf{p}}^n_g$ is the measured position of the GNSS phase center in the navigation frame, $\mathbf{t}^b_g$ is the IMU-GNSS lever-arm, $\mathbf{T}^n_w$ is a fixed world-to-navigation transformation obtained through the initial alignment, as  described in Sect. III-E.
	
	4) Wheel speed factor:
	\begin{equation}
		\mathbf{r}_s (\mathbf{x}_k) = \left( \mathbf{R}^{b_k}_{w} \mathbf{v}^w_{b_k} + \bm{\omega}^b_{wb} \times \mathbf{t}^b_s \right)_y - \tilde{v}_s
	\end{equation}
	where $\bm{\omega}^b_{wb}$ is the angular velocity of the IMU, $\mathbf{t}^b_s$ is the IMU-wheel lever-arm, $\tilde{v}_s$ is the measured wheel speed. Here we apply a simplified wheel speed model, while it is convenient to introduce other parameters like the scale factor.
	
	Besides, the marginalization \hl{factor is always taken into account, which contains the historical information. The generation of the marginalization factor is illustrated in Fig. 4. Note that the GNSS position factor and wheel speed factor related to the oldest states are also considered in the marginalization but are not depicted.}

	The generic design of the navigation-centered factor graph makes it flexible for different sensor schemes.
	Apart from the above factors, other common sensors, like GNSS doppler, magnetometer and specific motion models, could be easily added to the factor graph.
	In later discussions, we would pay most of the attention to the performance of a minimum monocular VIO system and a GNSS-integrated multi-sensor system for geo-related applications.

	\subsection{Other details}
	
	\begin{figure}[!t]
		\centering
		\includesvg[width=8.4cm]{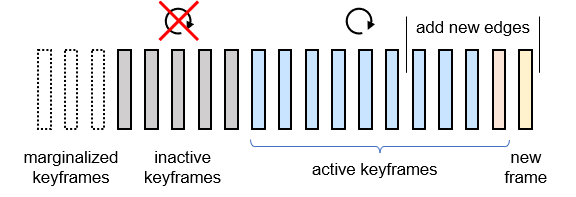}
		\caption{Illustration of  keyframe/edge management of DBA. New co-visible edges are appended within a small window and sparse mid-range edges are heuristically added. For mature edges that undergo enough iterations, the recurrent optical flow is no more performed.
		\hl{In our implementation, the maximum number of active edges (for optical flow update) is 48, the window size for DBA is 15, the range for co-visible edge appending is 5, the mid-range edges are tried between the newest frame and the frames  indexed at (-8,-9,-10)  relative to it.}
		}
		\label{fig_window}
	\end{figure}
	
	1) Keyframe and edge management
	
	In VO/VSLAM implementations, the selection and management of keyframes is important to maximize the estimation accuracy and reduce  computation costs. Besides, the dense-indirect BA structure of the system necessitates extra considerations for edge addition/deletion in the graph, which affects  the strength of visual association.
	
	\hl{In our system, everytime a new frame comes (termed as one new epoch), the co-visibility among frames is checked to append the new edges, and the update operation (optical flow + optimization) of the graph is performed  2 times.
		If the coming image is with enough disparity,  the second newest frame is retained  as the  new keyframe. In this case,  another iteration of update is performed, and the oldest keyframe is marginalized. Otherwise, the second newest frame is abandoned.}

	For edge management, we only append new co-visible edges within a small window of relatively new keyframes, as shown in Fig. \ref{fig_window}. In DROID-SLAM, a large window (25 by default) is used to check the co-visibility, thus to achieve better local consistency by introducing ``local loop closure". However, we find this uneconomical in our case and \hl{leads to high computational burden when the edges are dense.} It is found that, with the aid of IMU, a small window for edge appending is enough to maintain high-accuracy and locally consistent pose estimation. To maximize the accuracy, we still heuristically add mid-range edges with high co-visibility in the graph  and control the number of mid-range edges no more than 1 for a keyframe, which effectively reduces the \hl{peak computation load. The maximum number of the active edges (which need optical flow update) in the graph is 48, which are prioritized for relatively new edges.} For mature edges in the graph that undergo enough iterations, \hl{the optical flow is fixed while the re-projections are still  considered in DBA until marginalized.}

	2) Visual-inertial initialization
	
	As the IMU provides the awareness of scale and gravity direction, it is necessary to initialize the states in the metric-scale local frame \hl{for good convergence}. In this work, we simply use the initialization method proposed in VINS-Mono\cite{ref_vins}, where the  structure-from-motion is replaced by visual-only DBA. The DBA is iteratively performed based on the V-I initialization results, and the V-I translational extrinsic is taken into account after the first iteration. It is found that, with the accurate pose estimation given by DBA, the V-I initialization \hl{can} be finished with high precision under common maneuver. 
	
	3) GNSS integration
	
	As a global positioning technique, the position information provided by GNSS is expressed in earth-centered-earth-fixed (ECEF) frame. In this work, we simply align the trajectory produced by  VIO to the GNSS trajectory to estimate the 4-DOF world-to-navigation transformation $\mathbf{T}^n_w$ then just fix it
	\begin{equation}
		\hat{\mathbf{T}}^n_w = \mathop{\arg\min}\limits_{\mathbf{T}^n_w} \sum \left \| \mathbf{T}^n_w \circ \mathbf{T}^w_b \circ \mathbf{t}^b_g - \tilde{\mathbf{p}}^n_g \right \|_2
	\end{equation}
	where $\mathbf{T}^n_w$ comprises of a translation $\mathbf{t}^n_w$ and a heading $\theta^n_w$. The navigation frame is defined as the east-north-up frame anchored at a specific location, and the navigation-to-ECEF transformation $\mathbf{T}^e_n$ is known based on   WGS84.  \hl{The alignment is performed as long as the travelled distance in the window exceeds the threshold (e.g., 10 m).} 
	
	This strategy is simple and effective to integrate the GNSS information into the V-I system without introducing  a local-to-global transformation state. The uncertainty of the initial alignment is transferred to the pose uncertainty, which would be continuously refined considering the full observability of GNSS measurements.

	\section{Experiments}
	
	In this part,  we firstly test the monocular VIO pipeline of the system, termed as DBA-VIO, on the public TUM-VI and KITTI-360 datasets. After that, we test the multi-sensor integration performance based on the self-made \hl{vehicular dataset} collected in urban environments.  
	
	For DROID-SLAM, DBA-VIO and DBA-Fusion, we use images at  5 Hz and scaled to  $384\times 512$ (or equivalent level)  to guarantee real-time performance, which is verified sufficient for good results even in drastic motions. For navigation performance, we use real-time outputs of the system for evaluation, and the IMU-predicted poses are used for skipped frames if available. \hl{For dense mapping, the depth maps of marginalized keyframes are accumulated based on the estimated poses, and a multi-frame depth checking module\mbox{\cite{ref_droid}} is used to filter out dynamic objects and outliers.}

	\subsection{TUM-VI Dataset}
	\begin{figure}[!t]
		\centering
		\includesvg[width=8.4cm]{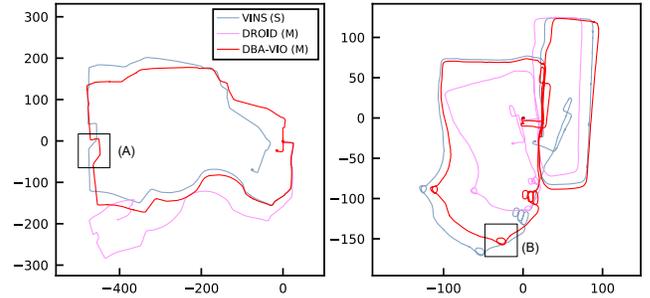}
		\caption{Trajectories of typical odometry schemes on ``outdoors1'' and ``outdoors7''. The trajectories are aligned with the ground-truth poses at the beginning of the sequence. The trajectory of DROID-SLAM (M) is scaled.}
		\label{fig_tum_vi_traj}
	\end{figure}
	\begin{figure}[!t]
		\centering
		\includesvg[width=7.0cm]{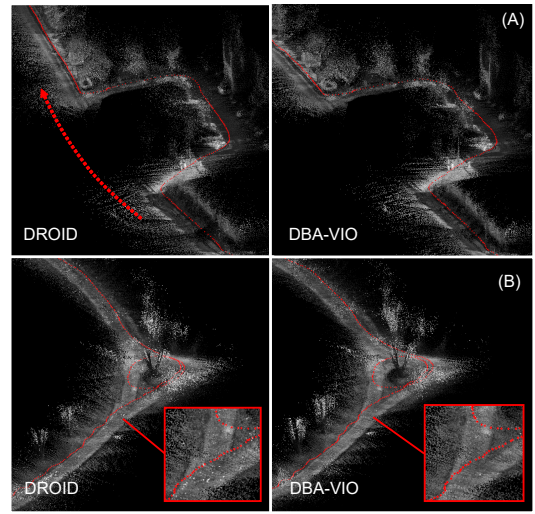}
		\caption{Dense mapping performance at the regions annotated in Fig. \ref{fig_tum_vi_traj}. For DROID-SLAM,  pointcloud bending and misalignment occur in (A) and (B) respectively, while DBA-VIO achieves good local map consistency.}
		\label{fig_tum_map}
	\end{figure}
	The TUM-VI benchmark\cite{ref_tum} is a visual-inertial dataset based on a handheld sensor platform, covering both indoor and outdoor scenarios. The  platform comprises a stereo camera and a low-cost BMI160 IMU. The ``magistrale" and ``outdoors" sequences are selected for evaluation due to the \hl{challenging} environment and relatively large scale. The ground truth poses from  motion capture system are available for the start and end segments of the sequences. On this basis, the absolute translation error (ATE) is calculated via the evo toolkit\cite{ref_evo}, which indicates the overall odometry drift in this case. 
	
	Apart from  DBA-VIO, different \hl{monocular VIO algorithms} are taken into account for comparison, including  VINS-Fusion (mono), \hl{ORB-SLAM3 (mono)} and DM-VIO (mono). Besides, we consider VINS-Fusion (stereo) and DROID-SLAM (mono) for extra reference. For \hl{ORB-SLAM3} and DROID-SLAM, we disable loop detection/global optimization to fairly evaluate the odometry accuracy.  For DROID-SLAM (mono) which is scaleless, we use the ground-truth poses available at the beginning to obtain a scale estimation, then use the scaled trajectory for global alignment to compute the ATE.
	
	The ATEs for different odometry schemes are listed in TABLE \ref{table_tumvi}. The proposed DBA-Fusion shows good performance on most of the sequences, which is more significant on the ``outdoor" squences. This could be attributed to the superiority of the deep optical flow to track environmental features even in low-texture environments, as well as the good compatibility of its iterative re-weighting mechanism in a VIO system. Besides, the trajectories of typical odometry schemes on ``outdoors1'' and ``outdoors7'' are shown in Fig. \ref{fig_tum_vi_traj}. The proposed method shows dramatically better translation and attitude estimation than the  visual-only DROID-SLAM, verifying the contribution of IMU integration to maintain low-drifting, metric-scale pose estimation. Compared to the VINS-Fusion (stereo) scheme, the proposed method obtains comparable scale estimation and shows better attitude maintenance. 
	\begin{table}[h]
		\caption{Absolute Translation Errors (m) \hl{of Different VIO Schemes} on TUM-VI Dataset.
		}
		\label{table_tumvi}
		\setlength\tabcolsep{4pt} 
		
		\begin{tabular}{l|llllll|l}
			\hline
			\makecell[c]{Seq.} &
			\makecell[c]{VINS\\stereo} &
			\makecell[c]{VINS\textsuperscript{1}\\mono} & 
			\makecell[c]{ORB\textsuperscript{2}\\mono} & 
			\makecell[c]{DMVIO\textsuperscript{1}\\mono}  & 
			\makecell[c]{DROID\textsuperscript{*}\\mono} & 
			\makecell[c]{DBA\\mono} & 
			\makecell[c]{Leng.\\ (m)}   \\ 
			\hline
			magi1 & \makecell[c]{2.82} & \makecell[c]{2.19} & \makecell[c]{5.65} & \makecell[c]{2.35} & \makecell[c]{5.28} & \makecell[c]{\textbf{1.47}} & \makecell[c]{918}\\
			magi2 & \makecell[c]{3.86} & \makecell[c]{3.11} & \makecell[c]{\textbf{0.72}} & \makecell[c]{2.24} & \makecell[c]{5.53} & \makecell[c]{1.81} & \makecell[c]{561}\\
			magi3 & \makecell[c]{3.44} & \makecell[c]{\textbf{0.40}} & \makecell[c]{4.89} & \makecell[c]{1.69} & \makecell[c]{38.82} & \makecell[c]{1.02} & \makecell[c]{566}\\
			magi4 & \makecell[c]{4.76} & \makecell[c]{5.12} & \makecell[c]{3.40} & \makecell[c]{1.02} & \makecell[c]{8.36} & \makecell[c]{\textbf{0.60}} & \makecell[c]{688}\\
			magi5 & \makecell[c]{1.79} & \makecell[c]{0.85} & \makecell[c]{2.94} & \makecell[c]{\textbf{0.73}} & \makecell[c]{6.72} & \makecell[c]{1.56} & \makecell[c]{458}\\
			magi6 & \makecell[c]{2.87} & \makecell[c]{2.29} & \makecell[c]{1.30} & \makecell[c]{1.19} & \makecell[c]{1.77} & \makecell[c]{\textbf{1.05}} & \makecell[c]{771}\\
			outd1 & \makecell[c]{49.76} & \makecell[c]{74.96} & \makecell[c]{70.79} & \makecell[c]{123.2} & \makecell[c]{220.0} & \makecell[c]{\textbf{13.85}} & \makecell[c]{2656}\\
			outd2 & \makecell[c]{17.10} & \makecell[c]{133.5} & \makecell[c]{14.98} & \makecell[c]{\textbf{12.76}} & \makecell[c]{60.55} & \makecell[c]{12.79} & \makecell[c]{1601}\\
			outd3 & \makecell[c]{\textbf{8.31}} & \makecell[c]{36.99} & \makecell[c]{39.63} & \makecell[c]{8.92} & \makecell[c]{44.06} & \makecell[c]{{8.68}} & \makecell[c]{1531}\\
			outd4 & \makecell[c]{13.74} & \makecell[c]{16.46} & \makecell[c]{25.26} & \makecell[c]{15.25} & \makecell[c]{26.12} & \makecell[c]{\textbf{4.45}} & \makecell[c]{928}\\
			outd5 & \makecell[c]{21.19} & \makecell[c]{130.6} & \makecell[c]{19.38} & \makecell[c]{7.16} & \makecell[c]{14.00} & \makecell[c]{\textbf{3.73}} & \makecell[c]{1168}\\
			outd6 & \makecell[c]{45.15} & \makecell[c]{133.6} & \makecell[c]{\textbf{16.84}} & \makecell[c]{34.86} & \makecell[c]{71.64} & \makecell[c]{19.55} & \makecell[c]{2045}\\
			outd7 & \makecell[c]{19.38} & \makecell[c]{21.90} & \makecell[c]{19.68} & \makecell[c]{5.00} & \makecell[c]{17.42} & \makecell[c]{\textbf{4.18}} & \makecell[c]{1748}\\
			outd8 & \makecell[c]{16.26} & \makecell[c]{83.36} & \makecell[c]{27.88} & \makecell[c]{\textbf{2.11}} & \makecell[c]{21.02} & \makecell[c]{7.25} & \makecell[c]{986}\\
			\hline
			ave(\%) & \makecell[c]{1.06} & \makecell[c]{3.20} & \makecell[c]{1.30} & \makecell[c]{0.85} & \makecell[c]{2.64} & \makecell[c]{\textbf{0.42}} & \makecell[c]{norm.}\\
			\hline
		\end{tabular}
		\begin{tablenotes}
			\item[*] Visual-only. \item[1] Statistics borrowed from\cite{ref_dmvio}.
			
			\item[2] Statistics partly borrowed from\cite{ref_orb}. Sequence magi[3,6], outd[1-4,6,8] which were reported loop closure are re-run with loop detection disabled.
		\end{tablenotes}
	\end{table}
	
	As to online dense mapping,  two scenes reconstructed by DROID-SLAM and DBA-VIO are shown in Fig. \ref{fig_tum_map}. For DROID-SLAM in online mode, the bending and misalignment of the pointclouds indicate significant pose drift, while this is almost overcome by DBA-VIO with the tight integration of a low-cost IMU. It is interesting that the low-drifting pose estimation of DBA-VIO has resulted in local consistent mapping performance without any loop closure techniques.

	\begin{figure}[!t]
		\centering
		\includesvg[width=8.4cm]{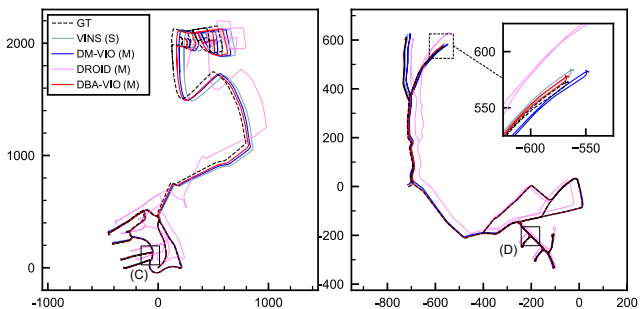}
		\caption{Trajectories of typical odometry schemes on ``0004'' and ``0005''. The trajectories are aligned with the ground-truth pose at t = 10 s. The trajectory of DROID-SLAM (M) is scaled.}
		\label{fig_0004_0005}
	\end{figure}
	\begin{figure}[!t]
		\centering
		\includesvg[width=7.0cm]{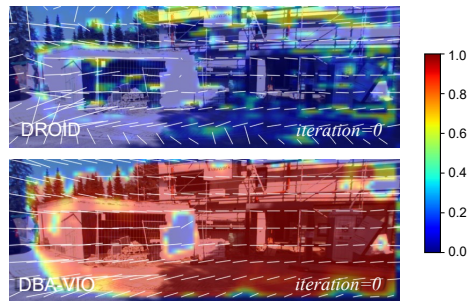}
		\caption{Optical flow produced by DROID-SLAM and DBA-VIO at the first iteration for one image pair in ``0002'' sequence of KITTI-360. The heat map indicates the weight (inverse uncertainty) of the optical flow.}
		\label{fig_weight}
	\end{figure}
	\begin{figure}[!t]
		\centering
		\includesvg[width=7.0cm]{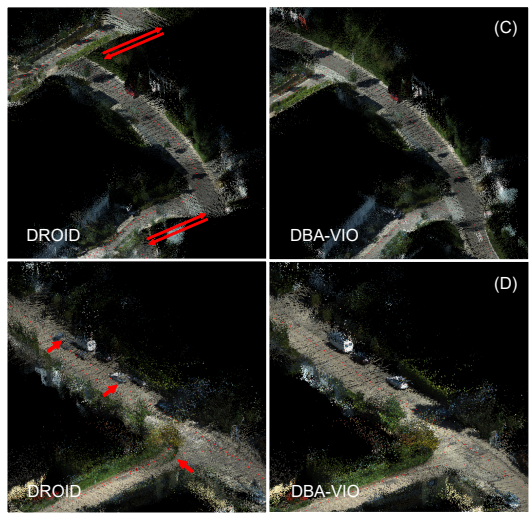}
		\caption{Dense mapping performance at the regions annotated in Fig. \ref{fig_0004_0005}. For DROID-SLAM,  severe misalignment and ghosting of the pointcloud is seen in (C) and (D) respectively.}
		\label{fig_kitti_map}
	\end{figure}
	\begin{table*}[h]
		\caption{Relative Pose Errors of \hl{Different VIO Schemes} on KITTI-360 Dataset. Best Monucular Results are \underline{Underlined} .}
		\label{table_kitti360}
		\setlength\tabcolsep{5pt} 
		\vspace{-10pt}
		\begin{center}
			\begin{tabular}{l|l|ll|ll|ll|ll|ll|ll}
				\hline
				\multirow{2}{*}{\makecell[c]{Seq.}} &
				\multirow{2}{*}{\makecell[c]{Desc.}} &
				\multicolumn{2}{c|}{\makecell[c]{VINS\\stereo}} & \multicolumn{2}{c|}{\makecell[c]{VINS\\mono}} & \multicolumn{2}{c|}{\makecell[c]{ORB\\mono}}   & \multicolumn{2}{c|}{\makecell[c]{DM-VIO\\mono}} & 
				\multicolumn{2}{c|}{\makecell[c]{DROID\textsuperscript{*}\\mono}}  & \multicolumn{2}{c}{\makecell[c]{DBA\\mono}} \\ \cline{3-14} 
				&
				&\makecell[c]{ $t_{rel}$\textsuperscript{1}} & \makecell[c]{$r_{rel}$\textsuperscript{1}} 
				&\makecell[c]{ $t_{rel}$} & \makecell[c]{$r_{rel}$} 
				&\makecell[c]{ $t_{rel}$} & \makecell[c]{$r_{rel}$}
				&\makecell[c]{ $t_{rel}$} & \makecell[c]{$r_{rel}$} 
				&\makecell[c]{ $t_{rel}$} & \makecell[c]{$r_{rel}$} 
				&\makecell[c]{ $t_{rel}$} & \makecell[c]{$r_{rel}$}   \\
				\hline
				0000 & Suburb & \makecell[c]{\textbf{0.626}} & \makecell[c]{0.212} & \makecell[c]{1.897} & \makecell[c]{0.176} & \makecell[c]{2.386} & \makecell[c]{0.117} & \makecell[c]{1.369} & \makecell[c]{0.129} & \makecell[c]{17.646} & \makecell[c]{0.205} & \makecell[c]{\uline{0.678}} & \makecell[c]{\textbf{0.105}}\\
				0002 & Suburb & \makecell[c]{0.591} & \makecell[c]{\textbf{0.169}} & \makecell[c]{1.006} & \makecell[c]{0.199} & \makecell[c]{1.309} & \makecell[c]{0.215} & \makecell[c]{0.724} & \makecell[c]{0.183} & \makecell[c]{8.031} & \makecell[c]{0.284} & \makecell[c]{\textbf{0.577}} & \makecell[c]{\uline{0.174}}\\
				0003 & Highway & \makecell[c]{\textbf{0.502}} & \makecell[c]{\textbf{0.073}} & \makecell[c]{2.754} & \makecell[c]{\uline{0.088}} & \makecell[c]{7.044} & \makecell[c]{0.151} & \makecell[c]{1.146} & \makecell[c]{0.111} & \makecell[c]{22.473} & \makecell[c]{0.144} & \makecell[c]{\uline{1.041}} & \makecell[c]{0.114}\\
				0004 & Suburb & \makecell[c]{0.692} & \makecell[c]{0.172} & \makecell[c]{1.710} & \makecell[c]{0.193} & \makecell[c]{1.976} & \makecell[c]{0.211} & \makecell[c]{1.063} & \makecell[c]{0.178} & \makecell[c]{13.636} & \makecell[c]{0.495} & \makecell[c]{\textbf{0.556}} & \makecell[c]{\textbf{0.153}}\\
				0005 & Suburb & \makecell[c]{\textbf{0.607}} & \makecell[c]{\textbf{0.207}} & \makecell[c]{1.187} & \makecell[c]{0.219} & \makecell[c]{1.414} & \makecell[c]{0.227} & \makecell[c]{0.729} & \makecell[c]{0.224} & \makecell[c]{6.969} & \makecell[c]{0.410} & \makecell[c]{\uline{0.619}} & \makecell[c]{\uline{0.209}}\\
				0006 & Suburb & \makecell[c]{\textbf{0.630}} & \makecell[c]{\textbf{0.149}} & \makecell[c]{1.349} & \makecell[c]{0.176} & \makecell[c]{1.685} & \makecell[c]{0.184} & \makecell[c]{0.887} & \makecell[c]{\uline{0.161}} & \makecell[c]{10.673} & \makecell[c]{0.209} & \makecell[c]{\uline{0.734}} & \makecell[c]{0.166}\\
				0009\,\textsuperscript{2} & Suburb & \makecell[c]{\textbf{0.486}} & \makecell[c]{\textbf{0.121}} & \makecell[c]{1.596} & \makecell[c]{0.144} & \makecell[c]{2.407} & \makecell[c]{0.184} & \makecell[c]{1.379} & \makecell[c]{0.136} & \makecell[c]{13.054} & \makecell[c]{0.275} & \makecell[c]{\uline{0.846}} & \makecell[c]{\uline{0.136}}\\
				0010 & Boulevard & \makecell[c]{\textbf{0.776}} & \makecell[c]{\textbf{0.206}} & \makecell[c]{3.610} & \makecell[c]{0.216} & \makecell[c]{5.335} & \makecell[c]{0.214} & \makecell[c]{2.130} & \makecell[c]{0.215} & \makecell[c]{20.522} & \makecell[c]{0.258} & \makecell[c]{\uline{1.486}} & \makecell[c]{\uline{0.208}}\\
				\hline
			\end{tabular}
		\end{center}
		\begin{tablenotes}
		\item[*] Visual-only. \item[1] $t_{rel}$ in \%, $r_{rel}$ in $^\circ$/100 m.	\item[2] Due to the gap of IMU data, only the first 1100 seconds are used. 
		\end{tablenotes}
	\end{table*}

	\subsection{KITTI-360 Dataset}
	The KITTI-360 benchmark\cite{ref_kitti360} is an autonomous driving dataset comprising multi-sensor data collected at the suburbs of Karlsruhe, Germany on 2013/5/28. The scenarios mainly include narrow suburban roads and highways, with relatively large scale and long durations. Among the mounted sensors, we use the perspective cameras and the OXTS IMU for the  test. The georegistered vehicle trajectory obtained by multi-sensor fusion is taken as the reference.
	
	Based on the high-precision ground-truth poses, we evaluate the odometry performance of different schemes (as mentioned in Sect. V-A) through relative pose errors, following\cite{ref_kitti}.  For the visual-only DROID-SLAM, we use the scaled trajectory obtained by whole trajectory $SIM(3)$ alignment. The statistics are listed in TABLE \ref{table_kitti360}. As shown in the table,   VINS-Fusion (stereo)  obtains the best performance, as the relatively long stereo baseline ($\approx$ 0.6 m) provides strong geometric information. Among the monocular schemes, the proposed DBA-VIO obtains superior performance, which is also indicated by the trajectories  in Fig. \ref{fig_0004_0005}. The improvement of introducing IMU into DROID-SLAM is dramatic, which maximizes the potential of the dense-indirect visual frontend by providing good robustness. As an indicator, Fig. \ref{fig_weight} shows the optical flow tracking performance with or without IMU when the vehicle turned left. With IMU aiding, the optical flow module \hl{can} generate good results even at the first iteration, thus contributes to better DBA convergence.
	
	We also show the online mapping results in Fig. \ref{fig_kitti_map}. The results are consistent with Sect. V-A, as the proposed DBA-VIO achieves locally consistent mapping performance based on the accurate pose estimation and significantly improves the practicability of DROID-SLAM in outdoor scenarios.
	
	\subsection{Self-Made Urban Dataset}
	
	To further evaluate the large-scale localization and mapping performance of the system, we turn to the self-made multi-sensor  dataset collected in Wuhan City on 2022/10/12. The experimental vehicle is equipped with RGB cameras, an ADIS16470 IMU and a Septentrio AsterRx4 GNSS receiver. A nearby base station is used for differential GNSS (DGNSS) processing. The smoothed trajectory of post-processing DGNSS/INS integration based on high-end IMU is taken as  reference. The reference solution is also used to simulate 1 Hz wheel speed measurements with 5 cm/s noise. \hl{The data sequence is  made publicly available.}
	
	\begin{figure}[!t]
		\centering
		\includesvg[width=8.0cm]{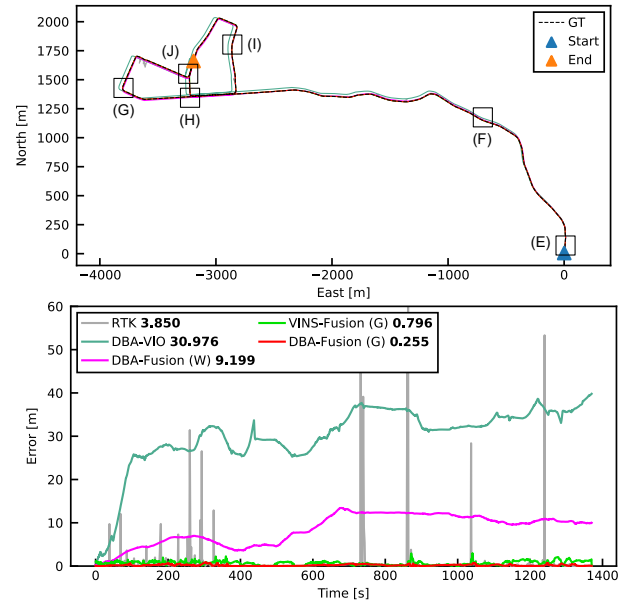}
		\caption{The trajectories and horizontal position error of different navigation schemes. The bold numbers are root mean square error (RMSE) in meters. The navgation scheme includes:  1) GNSS real-time kinematic (RTK); 2) VINS-Fusion with GNSS; 3) DBA-VIO; 4) DBA-Fusion with WSS; 5) DBA-Fusion with GNSS. All solutions are tested with \textbf{monocular}  setup.}
		\label{fig_whu_traj}
	\end{figure}
	\begin{figure}[!t]
		\centering
		\includegraphics[width=6.6cm]{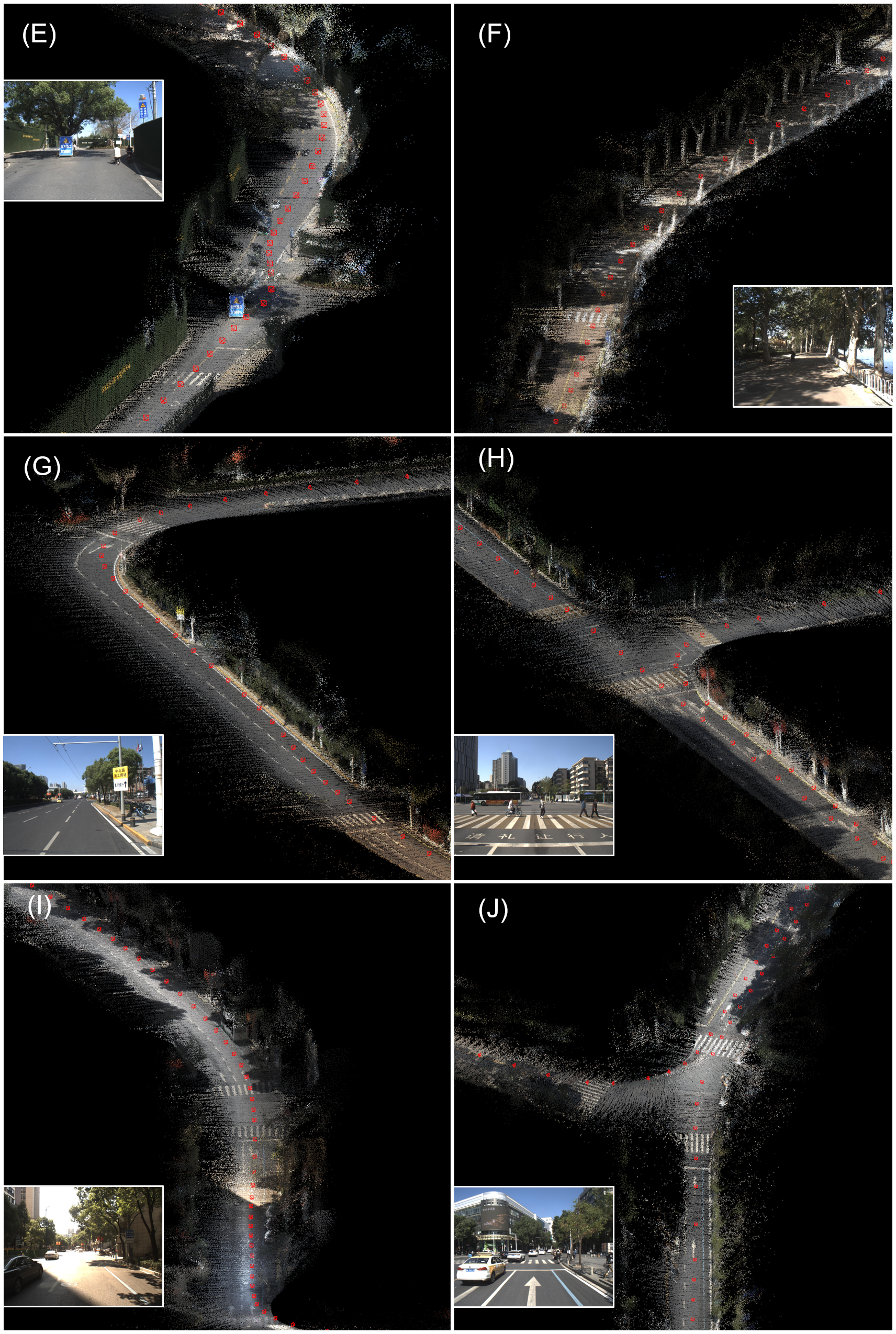}
		\caption{Mapping performance of DBA-Fusion (G) on the urban dataset.}
		\label{fig_real_map2}
	\end{figure}
	The trajectories and horizontal position errors of different navigation schemes are shown in Fig. \ref{fig_whu_traj}. As shown in the figure, the GNSS RTK scheme achieves driftless positioning but undergoes large errors during GNSS occlusion. For DBA-VIO, the position error drifts significantly at the beginning 100 seconds as the IMU excitation is insufficient. With wheel speed integration, the DBA-Fusion (W) scheme obtains smoother pose estimation performance with instant scale awareness. With GNSS RTK integration, the DBA-Fusion (G) scheme achieves stable decimeter-level position estimation. The result outperforms VINS-Fusion (G), which could be attributed to that the GNSS position information is tightly integrated and effectively constrains the VIO subsystem.
	
	Besides, the mapping performance of DBA-Fusion (G) is shown in Fig. \ref{fig_real_map2}. Generally, the proposed system could generate dense, accurate and georeferenced point clouds. As shown in \hl{Fig. 13 (H)} and (J), the mapping consistency is good when traversing the same area repeatedly, benefiting from the accurate global pose estimation. The dynamic objects are mostly filtered through multi-frame depth checking, leading to the black vacuum in Fig. 13 (J). This is one of the main limitations of the mapping functionality of the system, while this could be partly overcome  via multi-camera configuration. 
	
	In addition, the real-time performance of the system evaluated on Nvidia RTX 3090 GPU is shown in TABLE \ref{table_realtime}.
	
	\begin{table}
		\caption{Real-time performance of the proposed system.}
		\label{table_realtime}
		\begin{tabular}{l|l|lll} 
			\hline
			\multicolumn{2}{l|}{\multirow{2}{*}{Procedure}} & \multicolumn{3}{c}{Time cost (ms)}  \\ 
			\cline{3-5}
			\multicolumn{2}{l|}{}                           & TUM-VI & KITTI-360 & Wuhan          \\ 
			\hline
			\multicolumn{2}{l|}{Feature Extraction}         & 11.3    & 10.7       & 11.6            \\
			\multicolumn{2}{l|}{Graph Management \& Correlation}           & 27.2    & 32.3       & 25.3           \\
			\multirow{4}{*}{Iteration\textsuperscript{1}} & Optical Flow       & 21.3    & 21.2       & 20.0        \\
			& \hl{BA Hessian\textsuperscript{3}}         & \hl{3.6}    & \hl{4.0}       & \hl{3.6}            \\
			& \hl{Optimization\textsuperscript{3}}       & \hl{3.4}    & \hl{3.4}       & \hl{3.2}            \\
			& \hl{Depth Update\textsuperscript{3}}     & \hl{1.2}    & \hl{1.2}       & \hl{1.2}           \\
			\multicolumn{2}{l|}{Marginalization}            & 2.2    & 1.9       & 1.4            \\ 
			\hline
			\multicolumn{2}{l|}{Real-time Performance\textsuperscript{2}}                    & 1.47x   & 1.51x      & 1.62x           \\
			\hline
		\end{tabular}
		\begin{tablenotes}
			\item[1] \hl{Number of iterations is 2 for non-keyframe epoch, 3 for keyframe epoch (see Sect. III-D)} and 0 for skipped frames with little disparity.
			
			\item[2] Using 5 Hz images as mentioned above.
			
			\item[3] \hl{Total time of  2 low-level iterations (see Fig. 3).}
		\end{tablenotes}
	\end{table}
	
	\section{Conclusion}
	\hl{In this paper, we propose a framework that tightly integrates deep dense visual bundle adjustment with multiple sensors in a probabilistic manner, enabling  real-time localization and dense mapping. The experimental results validate the superior localization performance of the VIO implementation based on this framework, and show its flexibility to fuse GNSS, WSS for large-scale, georeferenced applications. }
	Future work will focus on extending
	this system to dynamic scenarios and neural map representations.

	\section*{Acknowledgments}
	The implemented DBA-Fusion is developed by the GREAT Group, School of Geodesy and Geomatics, Wuhan University. The numerical calculations
	in this paper have been done on the supercomputing system
	at the Supercomputing Center of Wuhan University.

\end{document}